%% file: access.tex
\documentclass{ieeeaccess}
\usepackage{cite}
\usepackage{amsmath,amssymb,amsfonts}
\usepackage{algorithmic}
\usepackage{graphicx}
\usepackage{textcomp}
\usepackage{kotex}
\usepackage{hyperref}
\DeclareMathOperator*{\argmin}{arg\,min}

\usepackage{bm}
\makeatletter
\AtBeginDocument{\DeclareMathVersion{bold}
\SetSymbolFont{operators}{bold}{T1}{times}{b}{n}
\SetSymbolFont{NewLetters}{bold}{T1}{times}{b}{it}
\SetMathAlphabet{\mathrm}{bold}{T1}{times}{b}{n}
\SetMathAlphabet{\mathit}{bold}{T1}{times}{b}{it}
\SetMathAlphabet{\mathbf}{bold}{T1}{times}{b}{n}
\SetMathAlphabet{\mathtt}{bold}{OT1}{pcr}{b}{n}
\SetSymbolFont{symbols}{bold}{OMS}{cmsy}{b}{n}
\renewcommand\boldmath{\@nomath\boldmath\mathversion{bold}}}

\newtheorem{theorem}{Theorem}

\newtheorem{assumption}{Assumption}
\newtheorem{corollary}{Corollary}

\makeatother

\def\BibTeX{{\rm B\kern-.05em{\sc i\kern-.025em b}\kern-.08em
    T\kern-.1667em\lower.7ex\hbox{E}\kern-.125emX}}

\begin{document}
\history{Date of publication xxxx 00, 0000, date of current version xxxx 00, 0000.}
\doi{10.1109/ACCESS.2024.0429000}

\title{Safety-Ensured Robotic Control Framework for Cutting Task Automation in Endoscopic Submucosal Dissection}
\author{\uppercase{Yitaek Kim}\authorrefmark{1}, \MakeUppercase{Iñigo Iturrate}\authorrefmark{1}, \uppercase{Christoffer Sloth}\authorrefmark{1}, and \uppercase{Hansoul Kim}\authorrefmark{2}}

\address[1]{SDU Robotics, the Maersk Mc-Kinney Moller Institute, University of Southern Denmark, Odense, Denmark (e-mail: \{yik, inju, chsl\}@mmmi.sdu.dk)}
\address[2]{Division of Mechanical System Engineering, Myongji University, Yongin-si, Gyeonggi-do, Republic of Korea (e-mail: hansoul@mju.ac.kr)}

\tfootnote{This work was supported by 2024 Research Fund of Myongji University and Fabrikant Vilhelm Pedersen og Hustrus Legat.}

\markboth
{Author \headeretal: Preparation of Papers for IEEE TRANSACTIONS and JOURNALS}
{Author \headeretal: Preparation of Papers for IEEE TRANSACTIONS and JOURNALS}

\corresp{Corresponding author: Hansoul Kim (e-mail: hansoul@mju.ac.kr).}

\begin{abstract}
There is growing interest in automating surgical tasks using robotic systems, such as endoscopy for treating gastrointestinal (GI) cancer. However, previous studies have primarily focused on detecting and analyzing objects or robots, with limited attention to ensuring safety, which is critical for clinical applications, where accidents can be caused by unsafe robot motions. In this study, we propose a new control framework that can formally ensure the safety of automating the cutting task in endoscopic submucosal dissection (ESD), a representative endoscopic surgical method for the treatment of early GI cancer, by using an endoscopic robot. The proposed framework utilizes Control Barrier Functions (CBFs) to accurately identify the boundaries of individual tumors, even in close proximity within the GI tract, ensuring precise treatment and removal while preserving the surrounding normal tissue. Additionally, by adopting a model-free control scheme, safety assurance is made possible even in endoscopic robotic systems where dynamic modeling is challenging. We demonstrate the proposed framework in a simulation-based experimental environment, where the tumors to be removed are close to each other, and show that the safety constraints are enforced. We show that the model-free CBF-based controlled robot eliminates one tumor completely without damaging it, while not invading another nearby tumor.
\end{abstract}


\titlepgskip=-21pt

\maketitle
\input{sections/introduction}
\input{sections/preliminary_knowledges}

\input{sections/proposed_method}
\input{sections/experimental_design}
\input{sections/results}
\input{sections/comprehensive_discussion}
\input{sections/conclusions}
\input{sections/appendices}

\end{document}

%% file: sections/introduction.tex
\section{Introduction}\label{sec:introduction}
Gastrointestinal (GI) cancers account for a significant proportion of all cancers, comprising approximately 26\% of global cancer cases \cite{bray2018global}. These cancers are highly dangerous due to their often aggressive progression. Therefore, early detection and treatment are critical, as survival rates decline sharply with later stages of the disease \cite{hamashima2015mortality}.

Endoscopy is the most common method for screening for GI cancer, and it also enables treatment of lesions through surgical tools \cite{hamashima2015mortality, dekker2018advances, bevan2018colorectal, schreuders2016advances}. However, conventional endoscopes are specialized for diagnosis, limiting their ability to perform advanced interventions. In particular, the heavy weight of the endoscope places both physical and mental burdens on medical doctors \cite{lee2010two, lee2019robotic}. Additionally, the steering of the bending section is not intuitive, as it bends in perpendicular directions while both knobs are rotated along the same axis \cite{ruiter2012design, lee2019robotic}. Therefore, a method to improve the convenience of endoscope manipulation is required.

Endoscopic surgery robot systems that can overcome the limitations of conventional endoscopes have gained attention as next-generation platforms. Unlike conventional endoscopes, robotic systems have improved task performance due to the more dexterous movements of the end-effector. Additionally, the operator can control the entire robot through a master controller, allowing for intuitive control and multitasking, which offers great potential for endoscopic surgery\cite{abbott2007design, lomanto2015flexible, chiu2021colonic, hwang2020evaluation, berthet20182, nageotte2020stras, nakadate2020surgical, hwang2020k, kim2023endoscopic}. However, due to various technical limitations, endoscopic surgical robots are still restricted to teleoperation control in clinical environments.

Recent advancements in artificial intelligence (AI) technologies have driven active research into methods for automating sub-tasks of surgical procedures using robotic systems in laboratory environments \cite{hwang2022automating, hwang2020superhuman, rosen2015autonomous, lum2008telerobotic, gonzalez2021deserts, xu2021surrol, seita2018fast, murali2015learning, dharmarajan2023automating, dharmarajan2023trimodal, dharmarajan2023robot}. Researchers have demonstrated promising results in automating tasks such as peg transfer \cite{hwang2022automating, hwang2020superhuman, rosen2015autonomous, lum2008telerobotic, gonzalez2021deserts, xu2021surrol}, debridement \cite{seita2018fast, murali2015learning}, and shunt insertion \cite{dharmarajan2023automating, dharmarajan2023trimodal, dharmarajan2023robot}. Although various AI-based methods that could assist in automating sub-tasks of surgical procedures have been proposed, the main reason they have not yet been applied in clinical environments is the absence of safety assurance. Ensuring safety in robotic automation is crucial \cite{ferraguti2022safety}, especially in complex and sensitive systems like surgical robots, where unintended collisions could occur. However, the methods proposed in previous studies have primarily focused on accurately detecting and analyzing the objects or robots being manipulated. To apply robots in actual clinical environments, it is also necessary to develop methods that can prevent accidents caused by unsafe robot movements.



One of the most widespread methods to ensure safety in the robot movement is virtual fixtures. Virtual fixture is a method that defines virtual boundaries or constraints to guide the movements of a robot along a specific trajectory or restrict its operation within a designated space \cite{bowyer2013active}. Especially in the field of surgical robotics, it has been widely applied to improve safety by preventing the robot from approaching sensitive organs or lesions \cite{attanasio2021autonomy}. By using both \textit{guidance} virtual fixtures and \textit{forbidden region} virtual fixtures together, the surgical robot can track an accurate trajectory while avoiding areas that should not be accessed. However, the mathematical expressions that define virtual fixture constraints typically exhibit discontinuities. This is an issue, because such discontinuous constraints can cause damage to the lesion tissue due to unintended movements, significantly impacting the quality and reliability of robotic surgery. Furthermore, virtual fixture-based approaches lack  formal safety guarantees, which might lead to serious safety violations.


\begin{figure}[t]
   \centering
   \includegraphics[width=0.5\textwidth]{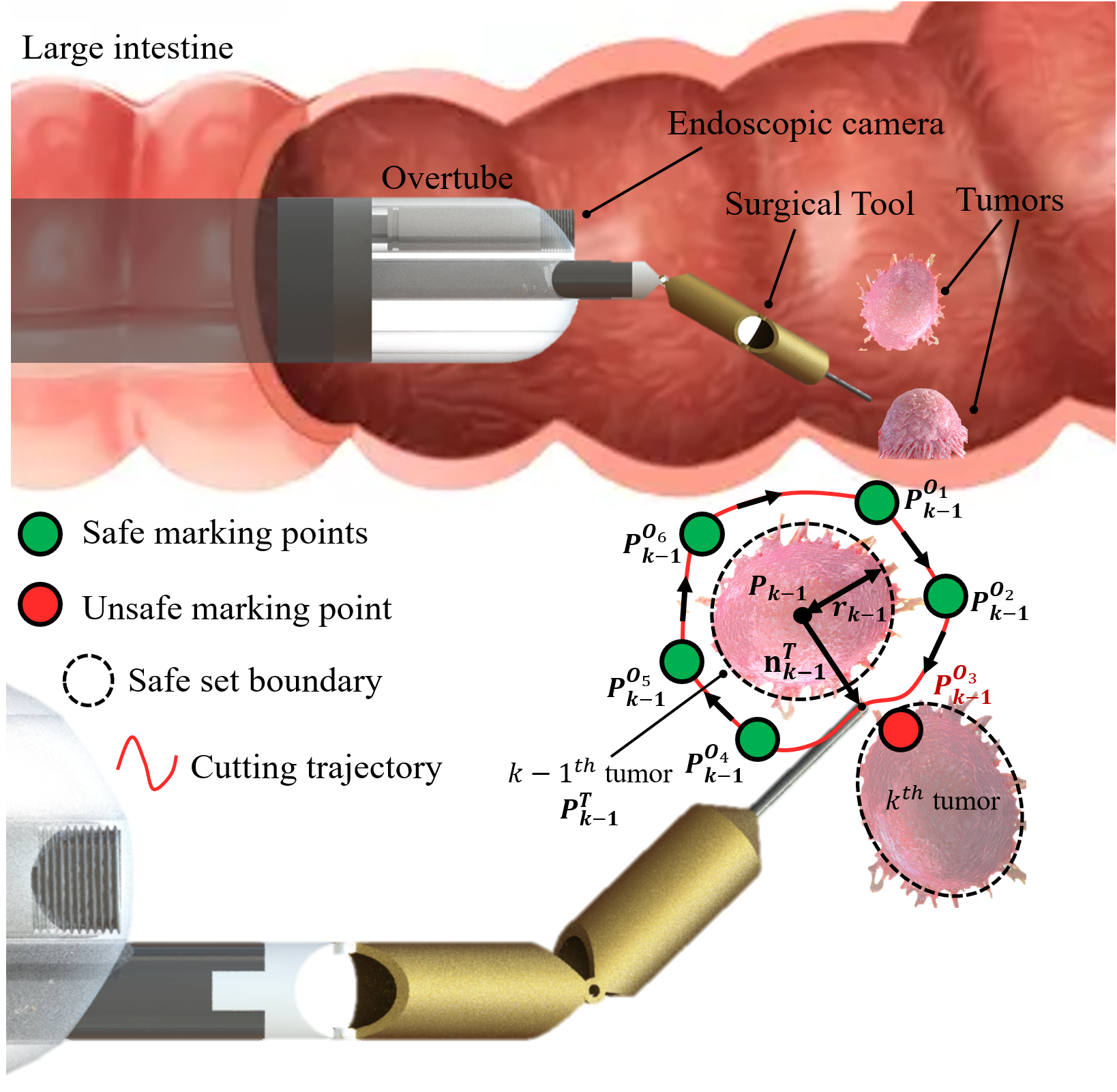} 
\vspace*{-3mm}    \caption{Conceptual illustration of the proposed approach for automated robotic motion in endoscopic surgery, enabling precise boundary distinction and individual tumor cutting while preserving normal mucosal areas within the GI tract.}
   \label{fig:overall_concept}
\end{figure}

To address the above-mentioned problem, a promising approach is to leverage Control Barrier Functions (CBFs), which can be used to ensure set invariance \cite{Ames2019CBFtheoryandapplications}. CBFs ensure forward invariance of a user-defined set and are efficiently integrated with quadratic programming. This allows CBFs to ensure safety across many safety-critical applications \cite{WOLF202218}\cite{Hadi2024}\cite{YKECC2024}. A practical challenge when using CBFs is that their safety guarantees rely on a system model that is is challenging to derive for robots that interact with human tissue.

To adapt this approach to robotic endoscopic surgery automation, one could consider recently-proposed safety-critical control designs based on robust and adaptive approaches\cite{lopez2020robust}\cite{JANKOVIC2018359} and data-driven methods \cite{ykgpracbf}\cite{Castaneda2021GPCBF}. On the other hand, it might be nontrivial to ensure safety guarantees for the complicated system dynamics of, for example, endoscopic robots based on tendon-driven mechanisms; thus, a control design in a model-free fashion could also be a potential way to achieve safety. Model-free safety critical control as proposed by \cite{molnar2021model} is specifically designed on a stable velocity controller in a robot, which is a firmly-established implementation in modern robotic systems \cite{Spong2005}. With a stable velocity controller, a control design with model-free CBFs provides a safe velocity control input to accomplish user-defined safety requirements without considering full-order robot dynamics. Furthermore, a control design based on model-free CBFs does not rely on a specific robotic application, so it can be efficiently used in general use cases. 

Based on Molnar's work \cite{molnar2021model}, we propose a new framework that can formally ensure the safety of certain processes involved in endoscopic submucosal dissection (ESD), a representative endoscopic surgical method for the treatment of early GI cancer \cite{bray2018global, hamashima2015mortality, oyama2005endoscopic}, by using an endoscopic robot (Fig. \ref{fig:overall_concept}). The purpose of ESD is not only to completely remove the tumor from the GI tract but also to ensure that the entire tumor is resected as a single piece, allowing for accurate pathological examination. In addition to the tumor boundaries, it is critical that the cutting depth is shallow enough to only pierce the submucosa, as too deep a cut could result in bleeding or puncturing of the intestine. In summary, as shown in Fig. \ref{fig:propsoed_framework}, the framework proposed in this study enables the automated motion of the robot to clearly respect the boundaries of each tumor, even when one or more tumors are located close to each other within the GI tract, while enforcing depth constraints. This allows for the individual removal of each tumor while preserving the normal mucosal areas between tumors.

The main contribution of this study can be summarized as follows: 


\begin{itemize}
    \item This paper proposes an integrated framework that ensures the safe automation of the cutting task in ESD by fusing clinical and engineering perspectives. The framework plans safe motions for tumor removal while enforcing safety boundaries as constraints to prevent unintended damage to surrounding tissue.
    \item This study extends the existing model-free CBF to medical robotic applications in where we consider clinical safety in the depth of the cut, contributing to prevent complications such as gastrointestinal perforation.
\end{itemize}



The remainder of this paper is organized as follows: Section \ref{sec:preliminary_knowledges} provides a detailed overview of control barrier functions and explains their relevance in ensuring safety in control systems. Section \ref{sec:proposed_method} describes the proposed framework, which includes a model-free safety-critical control scheme. Sections \ref{sec:experimental_Design} and \ref{sec:results} describe the simulation-based experimental setup and the results of the experiments. Sections \ref{sec:comprehensive_discussion} and \ref{sec:conclusion} present the discussion and conclusions of this study as well as an outline for further work.



%% file: sections/preliminary_knowledges.tex
\section{Preliminaries}\label{sec:preliminary_knowledges}
In this section, we revisit safety-critical control design. We present the concepts of Control Lyapunov Functions (CLFs) and Control Barrier Functions (CBFs). Subsequently, we introduce Input-to-State Stability (ISS) and Input-to-State Safety (ISSf), often used to deal with the impact of disturbances. Lastly, model-free CBFs are presented. 

\subsection{Stability with Control Lyapunov Functions}
Consider the following control affine system:
\begin{equation}
\dot{\bm{x}} = f(\bm{x}) + g(\bm{x})\bm{u}, \label{pre:sys_real}
\end{equation}
where $\bm{x} \in \mathcal{X} \subset \mathbb{R}^n$ is the system state, and $\bm{u}\in \mathcal{U}\subset\mathbb{R}^m$ represents a control input, and $f: \mathcal{X} \rightarrow \mathbb{R}^n$ and $g: \mathcal{X} \rightarrow \mathbb{R}^{n\times m}$ are locally Lipschitz continuous functions.We further assume that there exists a Lipschitz continuous controller, $k:\mathcal{X} \rightarrow \mathcal{U}.$
\begin{theorem}[\cite{Ames2019CBFtheoryandapplications}]
A function, $V: \mathcal{X} \rightarrow \mathbb{R}_{\geq 0}$ is a \textit{Control Lyapunov Function} (CLF) for \eqref{pre:sys_real} if there exist positive scalars, $c,k_1,k_2$ and a class $\mathcal{K}$ function, $\lambda$ such that:
\begin{align}
    k_1||\bm{x}||^c \leq &\text{ } V(\bm{x}) \leq k_2||\bm{x}||^c \\
    \inf_{\bm{u}\in\mathcal{U}}[L_{f}V(\bm{x}) + L_{g}V(\bm{x})\bm{u}] &\leq -\lambda\big(V(\bm{x})\big)\quad \forall\bm{x}\in \mathcal{X},
    \label{pre:clf_def}
\end{align}
where $L_{f}V(\bm{x}) + L_{g}V(\bm{x})\bm{u} = \frac{\partial V(\bm{x})}{\partial \bm{x}}\big(f(\bm{x}) + g(\bm{x})\bm{u}\big)$.
If $V$ is a CLF for system \eqref{pre:sys_real}, then any locally Lipschitz continuous controller, $\bm{u} = k(\bm{x})$ satisfying \eqref{pre:clf_def} ensures that $\bm{x} = 0$ is an exponentially stable equilibrium point of the system \eqref{pre:sys_real}.
\label{theo2}
\end{theorem}

\subsection{Safety with Control Barrier Functions}
Consider the system \eqref{pre:sys_real}, let $\mathcal{S}\subseteq \mathcal{X}$ 
 be the zero-superlevel set of  a continuously differentiable function, $h: \mathcal{X} \rightarrow \mathbb{R}$, i.e.:
\begin{equation}
    \mathcal{S} = \{\bm{x} \in \mathcal{X} \text{ }\vert\text{ } h(\bm{x}) \geq 0\}\label{pre:safe_set}.
\end{equation}
If the set, $\mathcal{S}$, is forward invariant, then the system \eqref{pre:sys_real} is safe with respect to $\mathcal{S}$ \cite{Ames2019CBFtheoryandapplications}. To ensure that $\mathcal{S}$ is forward invariant, the following theorem is considered:
\begin{theorem}[\cite{Ames2019CBFtheoryandapplications}]
A function, $h:\mathcal{X}\rightarrow \mathbb{R}$ is a \textit{Control Barrier Function} (CBF) for \eqref{pre:sys_real} if there exists an extended class $\mathcal{K}^{e}_{\infty}$ function $\alpha$ such that:
\begin{equation}
    \sup_{\bm{u}\in\mathcal{U}}[L_{f}h(\bm{x}) + L_{g}h(\bm{x})\bm{u}] \geq -\alpha\big(h(\bm{x})\big)\quad \forall\bm{x}\in \mathcal{S},
    \label{pre:cbf_def}
\end{equation}
where $\mathcal{S}$ is given in \eqref{pre:safe_set}.
If $h$ is a CBF for system \eqref{pre:sys_real}, then any locally Lipschitz continuous controller, $\bm{u} = k(\bm{x})$ satisfying \eqref{pre:cbf_def} ensures that $\mathcal{S}$ is forward invariant, which implies the system \eqref{pre:sys_real} is safe w.r.t  $\mathcal{S}$.
\label{theo1}
\end{theorem}

\subsection{Input-to-State Stability}
Consider the following system with disturbances:
\begin{equation}
    \dot{\bm{x}} = f(\bm{x}) + g(\bm{x})(\bm{u}+\bm{d}), \label{pre:dis_sys}
\end{equation}
where the bounded disturbance, $\bm{d}\in\mathbb{R}^m$, deteriorates the performance of CLF in terms of stability. To resolve the impact of disturbances, the concept of \textit{Input-to-State Stability} (ISS) \cite{sontagISS2008} can be introduced in \eqref{pre:clf_def} as follows:
\begin{equation}
    \inf_{\bm{u}\in\mathcal{U}}[L_{f}V(\bm{x}) + L_{g}V(\bm{x})(\bm{u}+\bm{d})] \leq  -\lambda\big(V(\bm{x})\big) +\iota(||\bm{d}||_{\infty}), \label{pre:iss_clf}
\end{equation}
where $\iota$ is an extended class $\mathcal{K}^{e}_{\infty}$ function; thus, we can achieve a robust stable controller satisfying \eqref{pre:iss_clf}.
It is worth noting that the concept allows the controller to converge to a neighborhood of the origin, is case of a nonzero disturbance. $\iota$ can be determined based on how conservatively the system should operate with respect to disturbances \cite{sontagISS2008}.

\subsection{Input-to-State Safety}
Similar to the ISS condition above, the safety condition in the presence of disturbances is extended to \textit{Input-to-State Safe} (ISSf) by defining the following larger safe set, $\mathcal{S}_{\delta} \supseteq \mathcal{S}$:
\begin{equation}
    \mathcal{S}_{\delta} = \{\bm{x} \in \mathcal{X} \text{ }\vert\text{ } h_{\delta}(\bm{x}) \geq 0\},
\end{equation}
where $h_{\delta}(\bm{x}) = h(\bm{x}) + \gamma(||\bm{d}||_{\infty})$ with class $\mathcal{K}$ function, $\gamma$. If there exists a locally Lipschitz continuous controller, $\bm{u}=k(\bm{x})$ satisfying the following constraint:
\begin{equation}
    L_{f}h(\bm{x}) + L_{g}h(\bm{x})(\bm{u}+\bm{d}) \geq  -\alpha\big(h(\bm{x})\big) -\gamma(||\bm{d}||_{\infty}), \label{pre:iss_cbf}
\end{equation}
then $\mathcal{S}_{\delta}$ is forward invariant and the system \eqref{pre:dis_sys} is ISSf \cite{ShishirISSfCBFs2019}.

\subsection{Model-Free Robotic Safety-Critical Control}
Instead of relying on the complex full-order dynamics, model-free safety-critical control design can be leveraged to ensure safety guarantees. Consider the following robotic system: 
\begin{equation}
M(\bm{q}) \ddot{\bm{q}} + C(\bm{q}, \dot{\bm{q}}) \dot{\bm{q}} + g(\bm{q}) = B\bm{u}, \label{pre:robot_dynamics}
\end{equation}
where $\bm{q} \in \mathbb{R}^n$ are configuration coordinates in configuration space, $\mathrm{Q}$, and $ M(\bm{q}) \in \mathbb{R}^{n \times n}$ is the inertia matrix,  $ C(\bm{q}, \dot{\bm{q}}) \in \mathbb{R}^{n \times n}$ represents the centrifugal and Coriolis forces, $g(\bm{q}) \in \mathbb{R}^n$ contains the gravity terms, and $B\in \mathbb{R}^{n\times m}$ is the input characteristic matrix.
Let us define a safe set with a continuously differentiable function, $h_q: \mathrm{Q} \rightarrow \mathbb{R}$ as follows: 
\begin{equation}
    \mathcal{S}_q = \{\bm{q} \in \mathrm{Q} \text{ }\vert\text{ } h_q(\bm{q}) \geq 0\}\label{pre:safe_set_confi}.
\end{equation}
And consider the following assumptions \cite{molnar2021model}:
\begin{assumption}
    The partial derivative of $h_q$ with respect to $\bm{q}$ is bounded such that $\big|\big|\frac{\partial h_q(\bm{q})}{\partial \bm{q}}\big|\big| \leq C_h$, $C_h>0$ for $\forall \bm{q} \in \mathcal{S}_q$. 
\end{assumption}
\begin{assumption}
    There exists the exponentially stable velocity controller, $\bm{u} = k_q(\bm{q},\dot{\bm{q}})$ where $k_q:\mathrm{Q}\times \mathbb{R}^n \rightarrow \mathcal{U}$, satisfying $||\dot{\bm{e}}(t)||\leq M||\dot{\bm{e}}_0||\mathrm{e}^{-\lambda t}$ for the positive scalar, $M, \lambda$, where $\dot{\bm{e}} = \dot{\bm{q}} - \dot{\bm{q}}_d$ is the tracking error between current velocity and desired one. In other words, a continuously differentiable Lyapunov function, $V(\bm{q},\dot{\bm{e}})$ exists such that:
    \begin{equation}
        k_1||\dot{\bm{e}}|| \leq V(\bm{q},\dot{\bm{e}}) \leq k_2||\dot{\bm{e}}||, \quad k_1, k_2 \in \mathbb{R}_{>0}
    \end{equation}
    and there exists $\bm{u}$ satisfying:
    \begin{equation}
        \dot{V}(\bm{q}, \dot{\bm{e}},\dot{\bm{q}}, \ddot{\bm{q}}_d, \bm{u}) \leq -\lambda V(\bm{q},\dot{\bm{e}}). \label{pre:stable_vel_controller}
    \end{equation}
\end{assumption}

Under these assumptions, a model-free safety controller that ensures safety guarantees can be designed based on the following theorem:
\begin{theorem}[\cite{molnar2021model}]
    Consider the system \eqref{pre:robot_dynamics}, the safe set, $\mathcal{S}_q$, safe velocity, $\dot{\bm{q}}_s$ ensuring 
        \begin{equation}
        \frac{\partial h_q(\bm{q})}{\partial \bm{q}}\dot{\bm{q}}_s \geq -\alpha\big(h_q(\bm{q})\big), \quad \forall\bm{q}\in \mathcal{S}_q, \label{pre:model_free_safety_condition}
    \end{equation}
    and a stable velocity tracking controller satisfying \eqref{pre:stable_vel_controller}. If the velocity controller tracks a safe velocity, $\dot{\bm{q}}_s$, fast enough \cite{molnar2021model},  and if $\lambda > \alpha$ such that the initial conditions, $(\bm{q}_0,\dot{\bm{e}}_0) \in \mathcal{S}_V$ can be defined as:
    \begin{equation}
        \mathcal{S}_V = \{(\bm{q},\dot{\bm{e}}) \in \mathrm{Q} \times \mathbb{R}^n \text{ }:\text{ } h_V(\bm{q},\dot{\bm{e}}) \geq 0\} \label{pre:init_condition_set},
    \end{equation}
    where $h_V(\bm{q},\dot{\bm{e}}) = -V(\bm{q},\dot{\bm{e}}) + \alpha_e h_q(\bm{q})$ with $\alpha_e = \frac{k_1(\lambda -\alpha)}{C_h}$, the system \eqref{pre:robot_dynamics} is safe with respect to $\mathcal{S}_q$.
\end{theorem}
It is worth noting that how fast the velocity controller tracks the safe velocity can be determined by using a Lyapunov certificate \cite{molnar2021model}\cite{Matni2024}. 


To deal with the impact of disturbances, we define a larger safe set, $\mathcal{S}_d \supseteq \mathcal{S}_q$, $\mathcal{S}_d = \{\bm{q} \in \mathrm{Q} \text{ }\vert\text{ }h_d(\bm{q})\geq 0\}$ where $ h_d(\bm{q}) = h_q(\bm{q}) + \gamma(||\bm{d}||_{\infty})$ and an inflated set corresponding to \eqref{pre:init_condition_set} as $\mathcal{S}_{V_d} = \{(\bm{q},\dot{\bm{e}}) \in \mathrm{Q} \times \mathbb{R}^n \text{ }:\text{ } h_{V_d}(\bm{q},\dot{\bm{e}}) \geq 0\}$ where $h_{V_d}(\bm{q},\dot{\bm{e}}) = h_V(\bm{q},\dot{\bm{e}}) +\gamma(||\bm{d}||_{\infty})$. The following corollary can be considered to ensure safety guarantees in the presence of disturbances:
\begin{corollary}[\cite{molnar2021model}]
    Consider the following system,
    \begin{equation}
M(\bm{q}) \ddot{\bm{q}} + C(\bm{q}, \dot{\bm{q}}) \dot{\bm{q}} + g(\bm{q}) = B(\bm{u}+\bm{d}), \label{pre:robot_dynamics_dis}
\end{equation}
the larger safe set, $\mathcal{S}_d$, and a velocity tracking controller satisfying ISS condition:
    \begin{equation}
        \dot{V}(\bm{q}, \dot{\bm{e}},\dot{\bm{q}}, \ddot{\bm{q}}_d, \bm{u},\bm{d}) \leq -\lambda V(\bm{q},\dot{\bm{e}}) +\iota(||\bm{d}||_{\infty}). \label{pre:ISS_vel_ctrl}
    \end{equation}
    If $(\bm{q}_0,\dot{\bm{e}}_0)\in \mathcal{S}_{V_d}$ and the velocity controller \eqref{pre:ISS_vel_ctrl} tracks a safe velocity, $\dot{\bm{q}}_s$ satisfying $ \frac{\partial h_d(\bm{q})}{\partial \bm{q}}\dot{\bm{q}}_s \geq -\alpha\big(h_d(\bm{q})\big), \forall\bm{q}\in \mathcal{S}_d$ and $\lambda > \alpha$, then we can say that the system \eqref{pre:robot_dynamics_dis} is ISSf with respect to $\mathcal{S}_d$.
    \label{pre:ISSf-model_free_cbf}
\end{corollary}

%% file: sections/proposed_method.tex
\begin{figure}
    \centering
    \includegraphics[width=\linewidth]{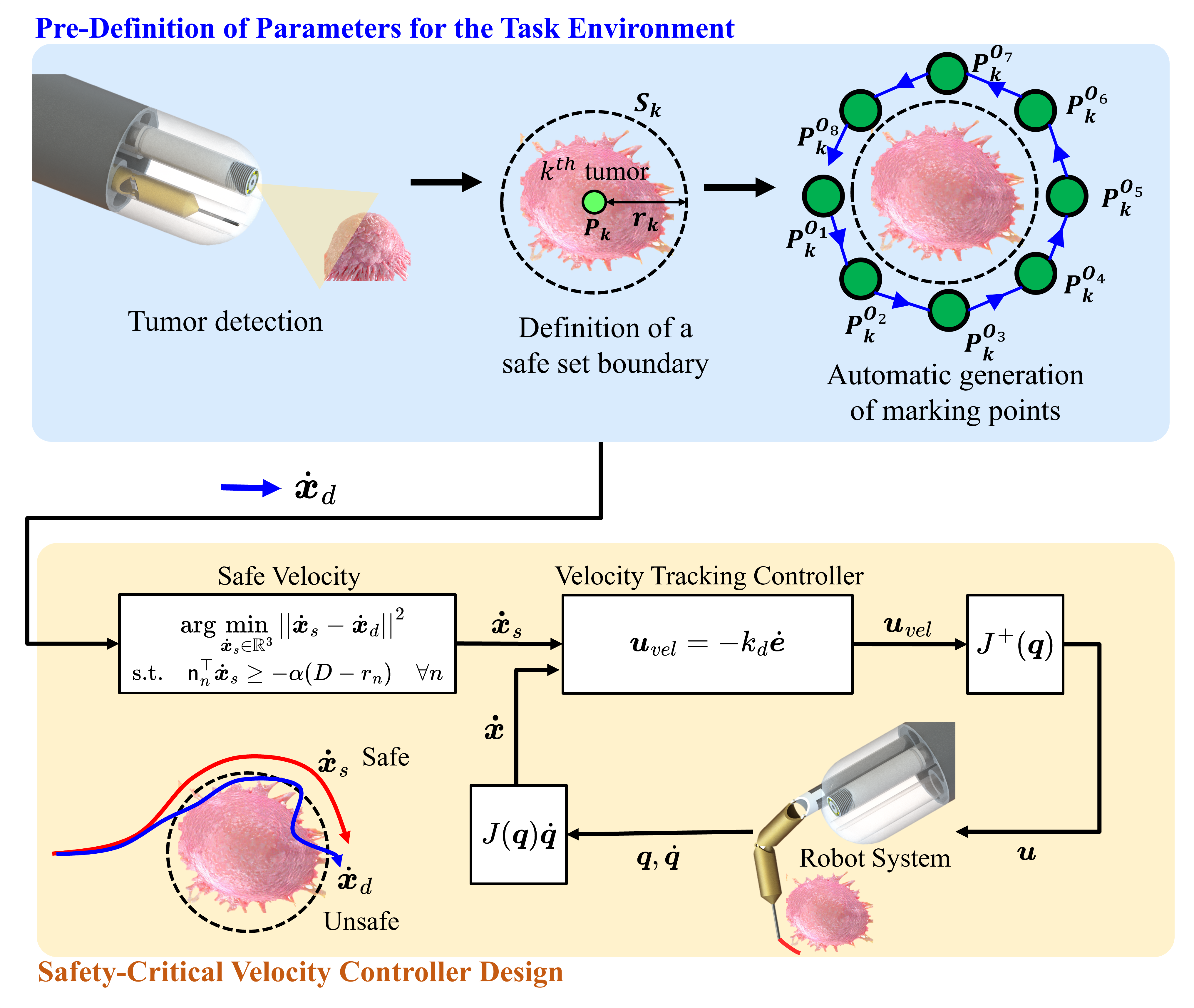}
    \caption{Illustration of the proposed framework for safe robotic ESD cutting task automation. Once the tumor is detected by the robot, the safe set boundary is defined to ensure safety. Then, a series of marking points are automatically generated around the tumor. Finally, cutting is performed along the trajectory defined by these marking points using a CBF-based safety-critical velocity controller.}
    \label{fig:propsoed_framework}
    \centering
    \vspace*{-3mm}
\end{figure}

\section{Application to Safety-Critical Robotic Cutting Task Automation in ESD}\label{sec:proposed_method}
Fig. \ref{fig:propsoed_framework} illustrates the framework we propose for ensuring formal safety in ESD cutting task automation. ESD can be subdivided into four main processes: injection, marking, cutting, and dissection \cite{lambin2021endoscopic}. Considering the difficulty of each process involved in ESD and the potential clinical safety issues, this study focuses on developing a method to ensure the safety of marking and cutting automation.





\subsection{Automatic Definition of Marking Points} \label{sec:auto_marking}
The marking process is to mark specific points around each tumor. Since the boundaries between the tumor and the surrounding tissue are often unclear, the marking process is essential for visually distinguishing the boundaries of the tumor \cite{lambin2021endoscopic}. This is also used to plan the cutting trajectory, so it is preferable to define the marking points in advance before the cutting process. Therefore, this process should be carried out even in the robotic ESD cutting task automation process to generate the primitive cutting motion trajectory and the safety of the endoscopic surgery.

As shown in Fig. \ref{fig:propsoed_framework}, the center position of the $n$-th tumor is represented as $P_{n} \in \mathbb{R}^3$ where $n = 1,2,\cdots,k$, and $k \in \mathbb{N}$ is the number of tumors. A set that includes the positions of multiple tumors required ESD by an endoscopic robot manipulator is defined as follows:
\begin{equation}
P = \{P_{1}, P_{2}, \cdots, P_{k}\} . \label{exp:tumors_set}
\end{equation}

Next, in order to perform the cutting process ideally in ESD, it is normally required to provide the reference trajectory that the robot tracks while performing the cutting process. To this end, we assume that the marking process to detect tumors and generate the marking points where the robot should pass has been completed previously, as shown in Fig. \ref{fig:propsoed_framework} \cite{kim2024efficacy}. Therefore, marking points, $P_{n}^{O_m}$, for each tumor, $P_{n}$, are generated around each center position of each tumor as follows:
\begin{equation}
P_{n}^{O} = \{P_{n}^{O_1}, P_{n}^{O_2},\cdots, P_{n}^{O_m}\}, \label{exp:tumors_marking_set}
\end{equation}
where $P_{n}^{O_m} \in \mathbb{R}^3$ represents the $m$-th marking point of the $n$-th tumor, $m \in \mathbb{R}$ is the number of marking points placed around the center position of each tumor. Consequently, the reference cutting trajectory (blue) as shown in Fig.~\ref{fig:propsoed_framework}, is provided by linearly interpolating all $P_{n}^{O_m}$ for each tumor.

In robotic ESD cutting task automation for tumor removal, safety may be compromised due to an abnormal reference trajectory for the robot or unexpected human-induced errors during teleoperation; thus, it is crucial to consider safety guarantees. For example, the control design for the surgery automation should prevent the endoscopic robot from damaging tumors. As damaging a tumor can alter the cells or tissues inside it, it is crucial to cut the tumor without causing any damage for accurate examination and pathological analysis. Furthermore, as too deep a cut can result in bleeding or damage to the intestine, it is critical for the control framework to also constrain the allowed depth.

To avoid damage to the tumors during the cutting procedure, we firstly define a safe set for each tumor in the following way:
\begin{equation}
    \mathcal{S}_n = \{\bm{x} \in \mathcal{X} \text{ }\vert\text{ } ||\bm{x}-P_n|| \geq r_n\},\label{pre:a_tumor_safe_set}
\end{equation}
where $\bm{x}$ is the position of the end-effector of the robot, and $r_n \in \mathbb{R}$ is the margin for ideal cutting of each tumor considering the tumor's size. Consequently, a safe set considering all tumors is represented by the intersection of safe sets, $\mathcal{S}_n$ as follows:
\begin{equation}
    \mathcal{S}^{*} = \mathcal{S}_1 \cap \mathcal{S}_2 \cap \cdots \cap \mathcal{S}_k,\label{pre:tumors_safe_set}
\end{equation}
which is assumed to be a non-empty set.

It should be noted that the boundary of the safe set $\mathcal{S}^{*}$ may not be smooth. To ensure safety, it is thus proposed to compute an inner-approximation $\mathcal{S}^{\circ}$ of $\mathcal{S}^{*}$, i.e. $\mathcal{S}^{\circ}\subseteq \mathcal{S}^{*}$. To enable this, $\mathcal{S}^{*}$ is restricted to be a compact semialgebraic set and $\mathcal{S}^{\circ}$ is defined using only one polynomial function. The computation of $\mathcal{S}^{\circ}$ can be accomplished using polynomial optimization, see Remark~4 in \cite{dabbene2017simple}.

\subsection{Robust Model-Free Safety Critical Control}
To ensure safety guarantees based on the safe set \eqref{pre:tumors_safe_set}, we leverage model-free CBFs \eqref{pre:model_free_safety_condition} to ensure that the position of the end-effector of the robot is inside $\mathcal{S}^{*}$. Consider the following candidate CBF inspired by \cite{molnar2021model} for each tumor from \eqref{pre:a_tumor_safe_set}:
\begin{equation}
h_n(\bm{x}) = D - r_n, \label{exp:a_tumor_cbf}
\end{equation}
where $D =  ||\bm{x}-P_n||$ is the distance between the current end-effector position and the center point of the tumor $P_n$. The gradient of the safe set, $\nabla h_n(\bm{x}) = \frac{(\bm{x}-P_n)^{\top}}{||\bm{x}-P_n||}$, is equal to the unit vector $\mathsf{n}_n^{\top}$, which points from the tumor center to the endoscopic robot. Lastly, the safe velocity controller used in \cite{molnar2021model} is designed by formulating the following quadratic program that minimally deviates from the desired velocity $\dot{\bm{x}}_{d}$:
\begin{align}
\argmin_{\dot{\bm{x}}_s \in \mathbb{R}^3}& || \dot{\bm{x}}_s - \dot{\bm{x}}_d||^2  \label{exp:qp}\\
\text{s.t.} \quad \mathsf{n}_{n}^{\top} \dot{\bm{x}}_s \geq &-\alpha (D-r_n) \quad \forall n, \nonumber
\end{align}
where $\dot{\bm{x}}_s$ represents the safe velocity that the robot should track to ensure safety based on the safe set. We choose the following velocity controller:
\begin{equation}
\bm{u}_{\textnormal{vel}} = -k_d\dot{\bm{e}} \label{exp:ctrl_law}
\end{equation}
where $\dot{\bm{e}} = J^{+}(\bm{q})(\dot{\bm{x}} - \dot{\bm{x}}_s)$, and $J^{+}(\bm{q})$ is the pseudo-inverse of the manipulator Jacobian matrix, $J(\bm{q})$, and $k_d\in \mathbb{R}_{\geq 0}$ is the controller gain. The control law, \eqref{exp:ctrl_law} does not include the model dynamics terms, $ M(\bm{q}), C(\bm{q}, \dot{\bm{q}}), g(\bm{q})$, but the controller satisfies the ISS condition from \eqref{pre:ISS_vel_ctrl}. From \textit{Corollary~\ref{pre:ISSf-model_free_cbf}}, we ensure that $\mathcal{S}^{*}$ is forward invariant with the velocity controller, \eqref{exp:ctrl_law}, tracking $\dot{\bm{x}}_s$, which ensures the ISSf condition with respect to $\mathcal{S}^{*}$. It is worth noting that, to the best of our knowledge, this is the first time that \textit{formal safety guarantees} in robotic ESD cutting task automation using a well-known model-free control law such as \eqref{exp:ctrl_law}, are explored.

%% file: sections/experimental_design.tex
\begin{figure}[t]
   \centering
   \includegraphics[width=0.5\textwidth]{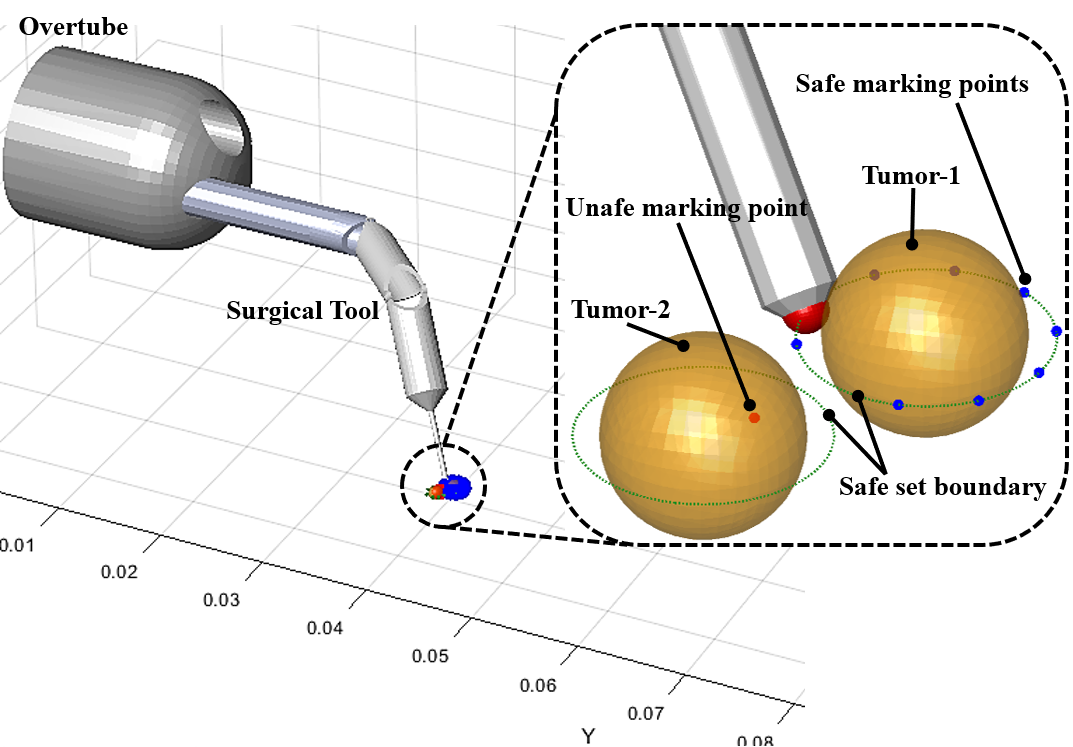} 
   \caption{The overall simulation environment for the experiments includes two tumors within the task environment of the robot. The safe set boundary for each tumor is defined by green dashed lines, blue circles represent safe marking points, and red circles indicate unsafe marking points.}
   \label{fig:exp_env}
   \vspace*{-3mm}
\end{figure}

\begin{figure}[t]
   \centering
   \includegraphics[width=0.5\textwidth]{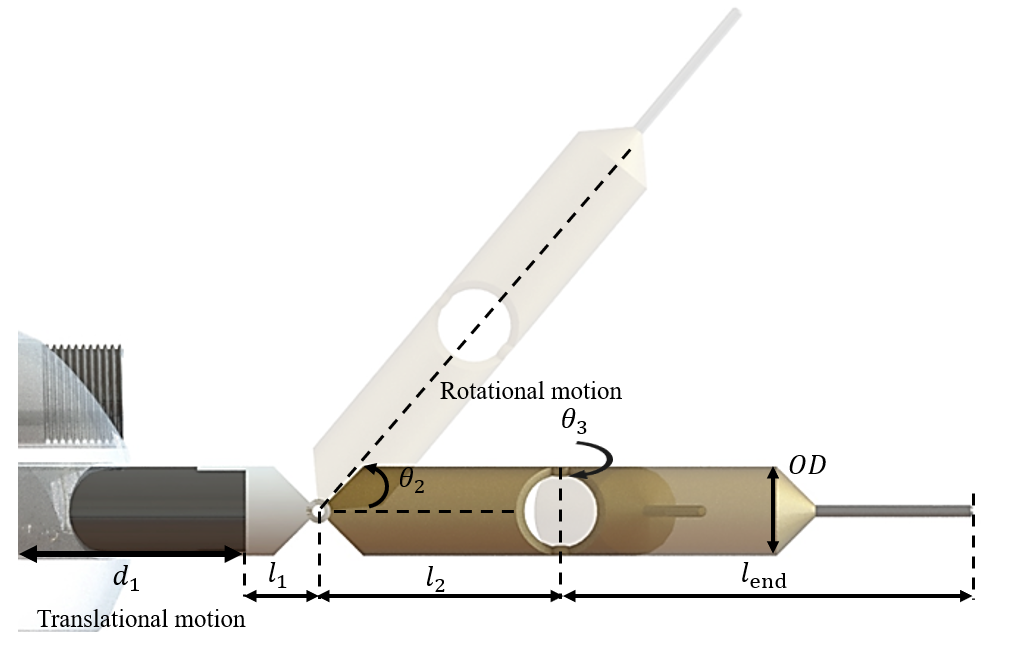} 
\vspace*{-3mm}    \caption{A three degree-of-freedom endoscopic robot manipulator designed to simulate automation for Endoscopic Submucosal Dissection in a three-dimensional space. The robot manipulator consists of a translational degree-of-freedom followed by two rotational degree-of-freedom.}
   \label{fig:mechanical_design}
\end{figure}
\begin{figure*}[t]
    \centering
    \includegraphics[width=0.8\textwidth]{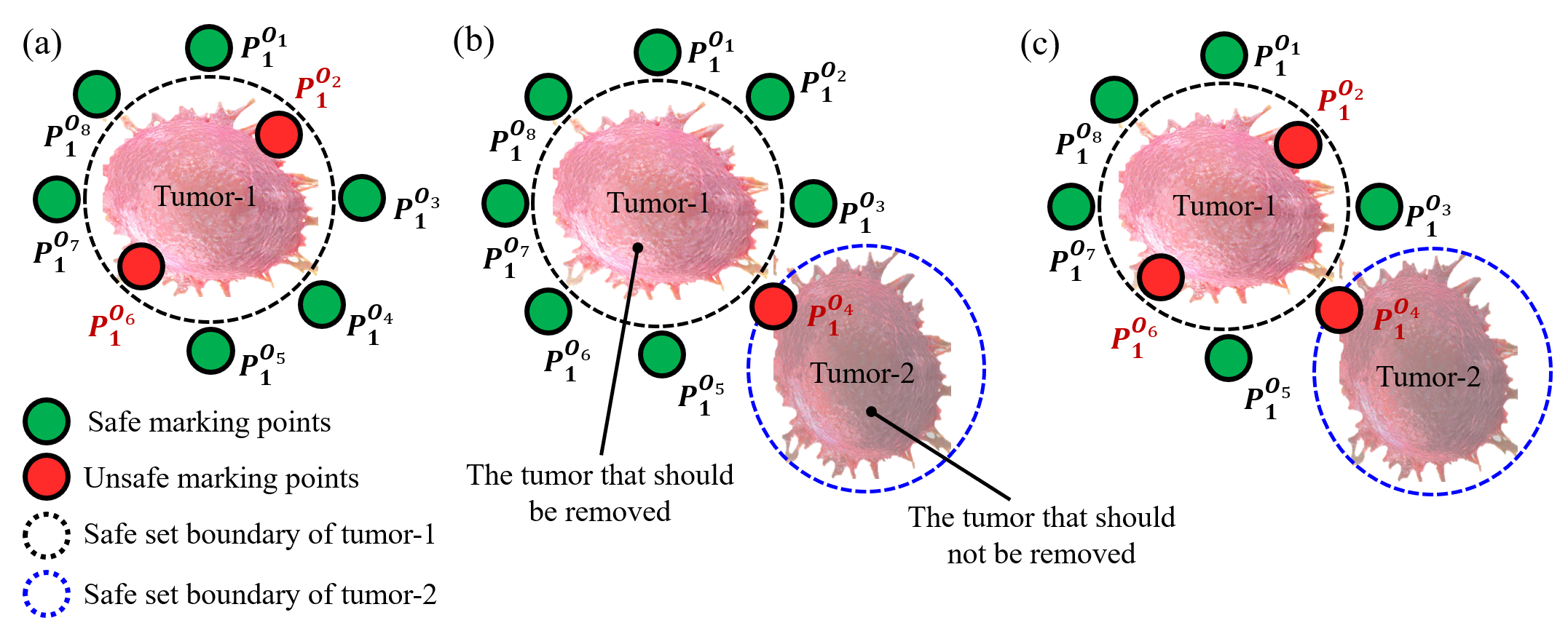}
    \caption{Three scenarios were considered in the experiment to verify the ensurance of theoretical safety: (a) A single tumor with two unsafe marking points inside the safe set boundary; (b) Two tumors in close proximity, with one unsafe marking point inside the safe set boundary of the tumor that should not be removed; and (c) Two tumors in close proximity, with two additional unsafe marking points inside the safe set boundary of the tumor that needs to be removed.}
    \label{fig:experiment_scenario}
\end{figure*}

\section{Simulation-Based Experimental Design}\label{sec:experimental_Design}
The main purpose of this experiment is to verify whether the method proposed in Section \ref{sec:proposed_method} is effective for application in ESD automation. Fig. ~\ref{fig:exp_env} provides an overview of the experimental environment. To focus on validating the theoretical effectiveness, this study performed experiments in a simulation environment. All simulation environments were set up in MATLAB.

\subsection{Mechanical Design and Numerical Simulation of an Endoscopic Robot Manipulator}\label{sec:numerical_simulation}

A three-degree-of-freedom (DOF) endoscopic robot manipulator was designed to perform cutting automation for ESD in a three-dimensional space (Fig. \ref{fig:mechanical_design}). This manipulator consists of one translational DOF and two rotational DOFs. To define the key parameters, such as the DOFs, outer diameter (OD), and total length, we referred to the geometric information of robots designed in previous studies on real endoscopic robot systems \cite{hwang2020evaluation, hwang2020k, kim2023endoscopic}. Based on these studies, the kinematic parameters of the endoscopic robot manipulator used in this experiment are as follows: $l_{1}$ = 3 mm, $l_{2}$ = 10 mm, $l_{end}$ = 17 mm, $OD$ = 3.7 mm.



To simulate the behavior of a robot, the dynamics of the robot are defined using \eqref{pre:robot_dynamics}. The torque, $\bm{\tau}$, applied to each joint of the robot serves as the system input. For Cartesian space control, the position of the robot end-effector can be calculated using the forward kinematics as follows:
\begin{align}
\bm{x} &= \Psi(\bm{q}),\label{exp:forward_kinematics}  \\
\bm{q} &= (d_{1},\theta_{2},\theta_{3}). \nonumber
\end{align}
Here, $\bm{x}$ represents the position of the robot end-effector, and $\Psi$ denotes the function representing the forward kinematics equations, and $l_1,l_2,l_{end}$ are the lengths of each link of the robot manipulator, and $d_{1},\theta_{2},\theta_{3}$ indicate the displacement and rotational motion of each joint, respectively. Since the method proposed in Section \ref{sec:proposed_method} is a velocity-tracking controller \eqref{exp:ctrl_law}, the error velocity, $\dot{\bm{e}}$ represents the error between the desired velocity $\dot{\bm{x}}_{s}$ -- which is the final output of the proposed method-- and the current linear velocity $\dot{\bm{x}}$ of the robot end-effector. Lastly, based on Equations \eqref{pre:robot_dynamics} and \eqref{exp:ctrl_law}, the state of the robot is updated in the simulation at each time based on: 
\begin{equation}
\ddot{\bm{q}}_{d} = M^{-1}(\bm{q})(\bm{u}_{\textnormal{vel}} - C(\bm{q},\dot{\bm{q}})\dot{\bm{q}} - G(\bm{q})).\label{exp:next_ddq1}
\end{equation}
In the following, we provide an overview of the experiment scenarios to verify the proposed framework.

\subsection{Experiment Scenarios}\label{sec:experiment_scenario}
Fig.~\ref{fig:experiment_scenario}  shows an overview of the experimental scenarios performed in this study. A maximum of two tumors exist within the task environment of the endoscopic robot, each was defined as either a tumor that needs to be removed or one that should be preserved. In this study, four scenarios were defined for this experiment as follows: (Scenario-1) A single tumor with two unsafe marking points inside the safe set boundary (Fig. \ref{fig:experiment_scenario}(a)); (Scenario-2) Two tumors in close proximity, with one unsafe marking point inside the safe set boundary of the tumor that should not be removed (Fig. \ref{fig:experiment_scenario}(b)); (Scenario-3) Two tumors in close proximity, with two additional unsafe marking points inside the safe set boundary of the tumor that needs to be removed (Fig. \ref{fig:experiment_scenario}(c)); (Scenario-4) An additional depth constraint for safety is considered above the third scenario (Fig.~\ref{sim:result_graph_ed}). The blue-filled circles represent ideal safe marking points, while the red-filled circles represent unsafe marking points that may pose clinical issues. Additionally, the green dashed circles represent the safe set boundary of each tumor. To generate the reference trajectory that robot should track, we linearly interpolate the marking points including safe and unsafe points. The final reference trajectory is colored as red dashed lines.

We evaluate the proposed framework under a couple of metrics including safety guarantees and conservatism in each scenario in the following section.

\begin{figure*}[t]
    \centering
    \includegraphics[width=1\linewidth]{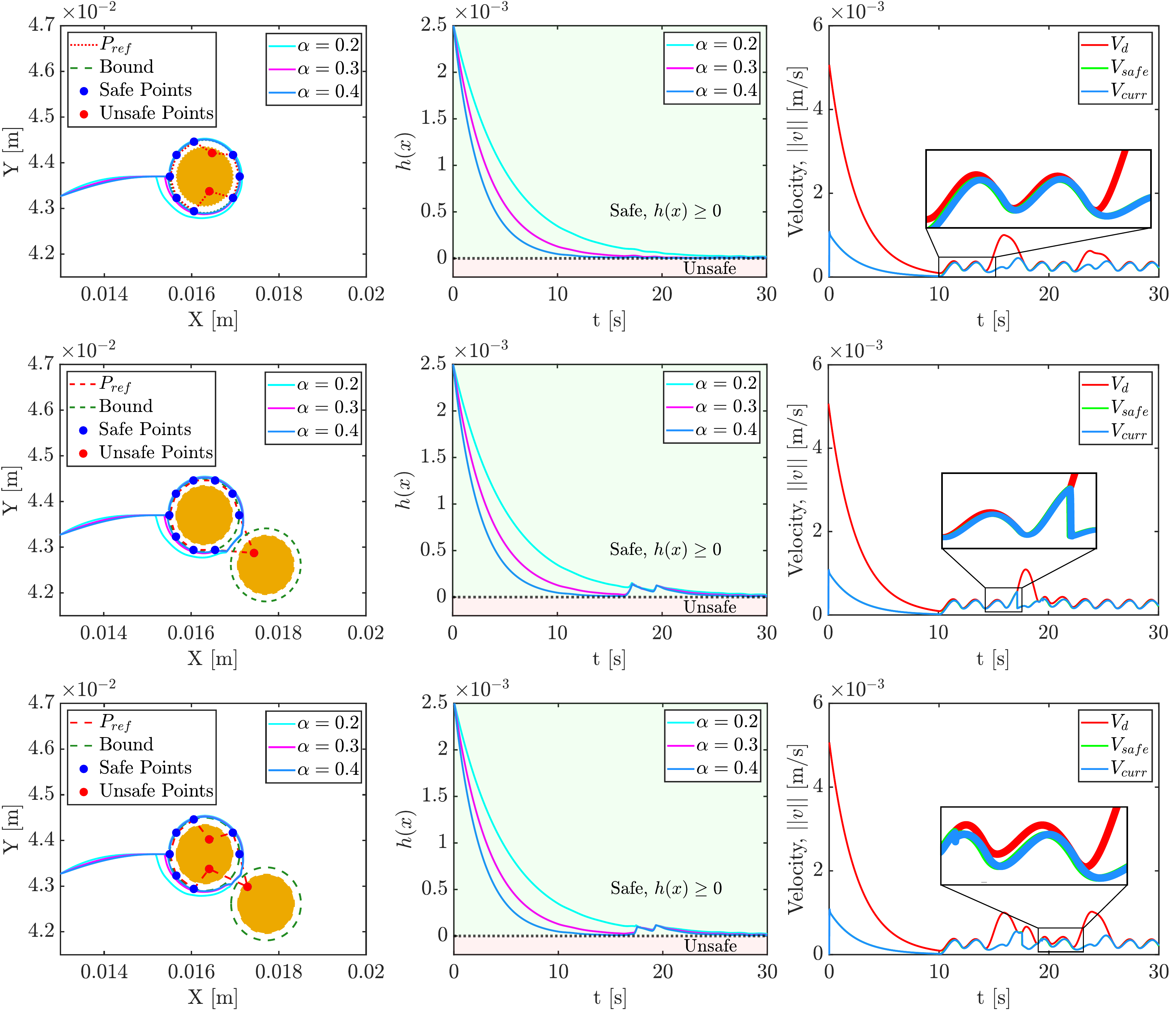}
    \vspace{-0.7cm}
    \caption{The simulation results of the proposed method. We conduct three scenarios with unsafe marking points. The rows show the results of each scenario and the three columns correspond to the position of the end-effector, safety guarantees with $h(\bm{x}) \geq 0 $, the performance of velocity tracking controller with $\alpha = 0.4$, respectively. Note that $V_d$, $V_{safe}$, and $V_{curr}$ are the desired, safe, and current velocities, respectively.}
    \vspace{-0.5cm}
    \label{sim:result_graph}
\end{figure*}

%% file: sections/results.tex
\section{Results}\label{sec:results}

\begin{figure*}[ht]
    \centering
    \includegraphics[width=1\linewidth]{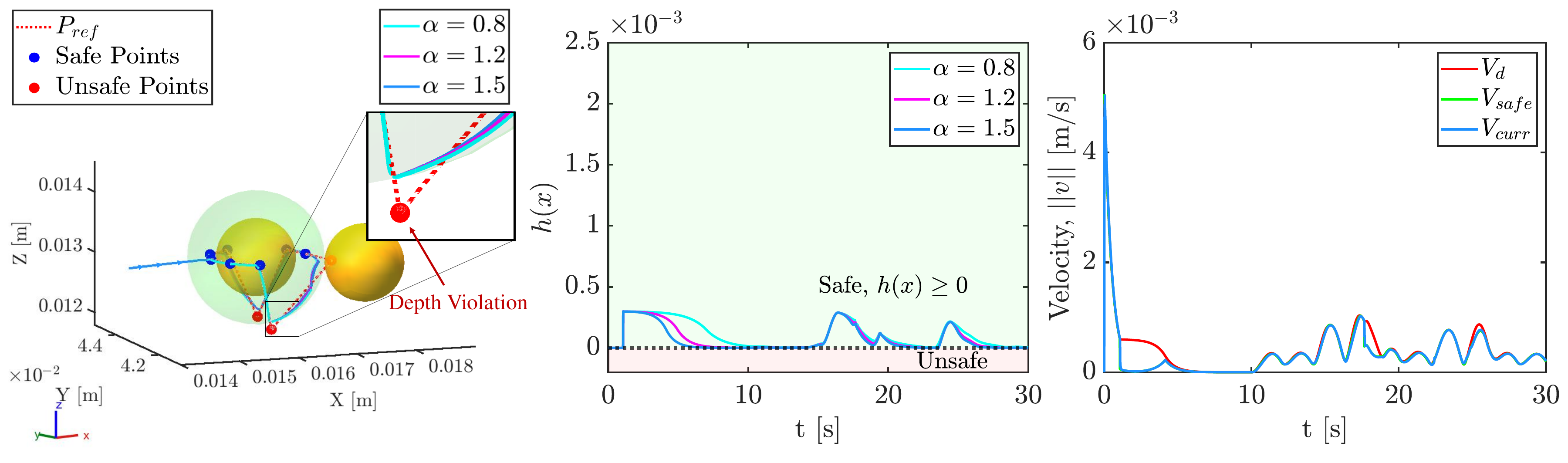}
    \vspace{-0.7cm}
    \caption{The simulation results of the scenario where we take into account depth violations. We use the same reference trajectory as what used in the third scenario. The first graph shows the whole position trajectory of the end-effector depending on $\alpha$. The second shows safety guarantees with $h(\bm{x}) \geq 0 $, and lastly the safe velocity tracking with $\alpha = 1.5$ is shown in the third graph. Note that the safe controller is activated after the robot reaches the safe area, which happens about $1$ second.}
    \vspace{-0.5cm}
    \label{sim:result_graph_ed}
\end{figure*}

In this section, we analyze the effectiveness of the proposed method under the following evaluation aspects at each scenarios: (1) Safety guarantees; (2) The level of conservatism based on the parameter, $\alpha$ from \eqref{pre:model_free_safety_condition}; (3) A stable safe velocity controller. Fig.~\ref{sim:result_graph} shows the simulation results in each scenario from Fig.~\ref{fig:experiment_scenario}. The rows show all results in the three scenarios. The first column shows marking points and the reference and the current position trajectories of the robot end-effector. The second column describes safety guarantees and the level of conservatism depending on $\alpha$. The last column shows the performance of a stable velocity controller.

In the first scenario, as shown in the first row of Fig.~\ref{sim:result_graph}, the end-effector position is trying to track the reference trajectory, $P_{ref}$ defined by the marking points. Even if the reference trajectory would cause the robot to cross into the tumor due to the unsafe marking points (red circle), the robot still achieves safe cutting motions without crossing the safety boundary (dashed green line) with the help of the CBF-based safe controller \eqref{exp:qp}. It is clearly observed that the safety guarantees are enforced at all times as shown in the safe constraint function, $h(\bm{x})$ of Fig.~\ref{sim:result_graph}. 

Furthermore, several simulations are carried out by varying the parameter, $\alpha$, in order to evaluate the level of conservatism. As shown in the middle graph of the first row of Fig.~\ref{sim:result_graph}, it is observed that the conservatism decreases as the alpha value increases. Note, however, that, as will be discussed in the following section, the safety guarantees might be violated if we choose an excessively large alpha \cite{molnar2021model}. Next, it is also shown that the velocity controller accurately tracks the safe velocity provided by the CBF-based safe controller \eqref{exp:qp} in the rightmost graph in the first row. It is worth noting that the CBF-based safe controller \eqref{exp:qp} provides the control input that is closest to the desired velocity reference, while at the same time ensuring safety constraints.

In the graphs of the second and third rows, the safety guarantees with the proposed framework are still achieved like the first scenario, although there exist unsafe marking points in the tumors as shown in the second and third rows of Fig.~\ref{sim:result_graph}. Depending on $\alpha$, similar conservative behavior to that observed in the first scenario can be seen in these two scenarios, as shown in the middle graphs of the second and third rows. Likewise, it is observed that the controller tracks the safe velocities in these two scenarios with results similar to the first scenario. 

To demonstrate that our proposed framework can also be used ensure that the depth of the cut is kept shallow enough to avoid damage to the intestine or submucosal tissue, we conduct an additional scenario where we include depth constraint while cutting a tumor and also compare the results with different parameters, $\alpha$. As shown in Fig. 7, we defined a larger outer constraint sphere covering the tumor, colored pale green, in order to consider cutting depth. Subsequently, we designed the safe set between the inner sphere (a tumor to be removed) and the outer one while using \eqref{exp:a_tumor_cbf} and the control barrier function as defined by $h_n(\bm{x}) = -D + r_n \geq 0$. As discussed in Molnar et al. \cite{molnar2021model}, the closest control barrier function is considered to provide the safe velocity reference in equation (23). For example, when the robot is about to enter the tumor, the direction of the safe velocity, $\frac{(\bm{x}-P_n)^{\top}}{||\bm{x}-P_n||}$ should be outward. Conversely, when the robot approaches the boundary of the outer sphere, the direction of the safe velocity should be inward. As a result, the safe cutting trajectory is also ensured with depth constraint as shown in the first and second graphs of Fig.~\ref{sim:result_graph_ed}, while showing that $h(\bm{x})$ is always over $0$. Likewise, it is found that the larger $\alpha$ reduces the conservatism of the controller as well. Compared to the previous scenarios, we choose a higher $\alpha$ value since the current case has the smaller safe region, and the less conservative performance of the safe controller is necessary to ensure the safety guarantees with the region. Note that $\alpha$ values used in the simulations were arbitrarily selected to ensure the safety guarantees. Tuning the values requires additional analysis, which is beyond the scope of this paper; we will briefly discuss this aspect in the following section.


%% file: sections/comprehensive_discussion.tex
\section{Discussion}\label{sec:comprehensive_discussion}
This study proposed a novel framework that can ensure the safety of automation for robotic surgery in a theoretically rigorous way. Our results were demonstrated in a scenario based on ESD, a representative endoscopic surgical method for the treatment of early-stage GI cancer, by using an endoscopic robot. The proposed framework utilizes CBFs to enable the automated motion of the robot, allowing it to clearly respect the boundaries of each tumor, even when multiple tumors are in close proximity within the GI tract. This allows individual removal of each tumor while enforcing safety guarantees on the surrounding normal mucosal tissues.

Although this study focuses on ensuring the safety of cutting task automation during ESD, extending the application to various tasks such as suturing or clamping by including not only the definition of the safe set boundary but also a model that accounts for the nonlinear interactions between superelastic tissues and the robot is required to bring the framework closer to practical clinical application. Additionally, considering superelastic characteristics, such as the texture and stiffness of the tumor, are considered, is needed to validate the robustness of the framework to variability that may arise in real-world applications.

This study performed all experiments in a high-fidelity simulation environment, where the system model is accurately known, to primarily verify the feasibility and validity of the proposed framework. As shown in Fig. \ref{sim:result_graph}, even when multiple tumors are located close to each other and unsafe marking points are defined, it was confirmed that the robot can automatically remove the target tumor without damaging the surrounding tissues by enforcing the constraints of the safe set boundary for each tumor. 

A model-free control scheme was leveraged, thus avoiding the complexity of robot dynamics modeling. However, even though a model-free control scheme was used, the simulation environment can differ from the actual environment where the robotic system is implemented. From \eqref{exp:ctrl_law}, the proposed framework employs a simple model-free velocity controller to track the safe velocity $\dot{\bm{x}}_s$. Here, since $\dot{\bm{e}}$ is expressed as $J^{+}(\bm{q})(\dot{\bm{x}} - \dot{\bm{x}}_s)$, it is crucial to accurately measure the linear velocity of the end-effector of the robot in real time. Without additional sensors on the end-effector, the linear velocity of the end-effector can be calculated by using the joint velocities in the joint space and the Jacobian matrix. However, in tendon-driven mechanisms like endoscopic robots, it is not possible to attach sensors such as encoders to each joint. Moreover, due to the hysteresis problem \cite{kim2020effect}, the commanded motion at the motor may not be perfectly applied to each joint of the robot, which can lead to low accuracy in the calculated linear velocity of the end-effector. Therefore, by integrating a method capable of overcoming hysteresis with nonlinear characteristics, such as the Bouc-Wen model \cite{do2014hysteresis}, for estimating the linear velocity of the end-effector, ideal results similar to those in the simulation environment are expected to be achieved in the actual operating environment of the robotic system. As an alternative to hysteresis modeling, a system with two motors per bending degree of freedom that mechanically compensates for hysteresis by minimizing slack could be adopted \cite{le2016survey}.

As explained in Section \ref{sec:auto_marking}, a marking process should be performed before the cutting process for tumor removal. As indicated by \eqref{exp:tumors_set} -- \eqref{exp:a_tumor_cbf}, this study assumes that all information, such as marking points, tumor sizes, and locations, is provided beforehand. However, in actual clinical environments, the geometric information and locations of the tumors should be estimated from the images inputted from the endoscopic camera mounted on the robot system, or registered during the pre-operative examination stage. To enable the proposed method to be used in real clinical environments, additional algorithms need to be integrated into the framework. For example, using stereo matching algorithms like RAFT-Stereo \cite{lipson2021raft} and segmentation algorithms such as the MedSAM \cite{ma2024segment} would allow for three-dimensional analysis of objects in the task environment. This approach would enable the estimation of tumor shapes, sizes, and locations. The method proposed in this study focuses on ensuring the safety of automated robotic motion. When combined with the aforementioned deep learning-based vision algorithms for precise tumor localization, the proposed framework is expected to contribute to the ongoing advancement of medical robotics.

When it comes to the model-free control barrier functions used in the proposed framework, $\lambda$ in the velocity controller should be properly decided in any controllers to design the safe controller \eqref{pre:model_free_safety_condition}. In the simulations, we arbitrarily chose the value to implement a stable velocity controller in a heuristic manner. To determine $\lambda$ in a constructive way, we can estimate the parameter based on learning approaches from data based on contraction metrics and Lyapunov analysis \cite{pmlr-v155-boffi21a}. 

Additionally, unexpected external forces in the real-world experiment could happen due to the uncertainty of the system model such as the robot model, deformability of the body tissues (e.g. GI), and the interaction forces between the robot and tissues, which could lead to safety violations. However, the unexpected external forces could be dealt with as one of the disturbances $\bm{d}$ in the controller as represented in \eqref{pre:robot_dynamics_dis}. Although we did not consider those uncertainties when cutting tumors in the simulations conducted in this paper, it is possible to achieve robust safety against unexpected external forces arising from uncertainties, since we use the robust CBF-based controller with input-to-state stability and safety conditions, based on \textit{Corollary}~\ref{pre:ISSf-model_free_cbf}. Alternatively, we can ensure safety guarantees while reducing unknown uncertainty simultaneously by estimating parametric and non-parametric uncertainty based on adaptive CBF\cite{Taylor2020}, robust adaptive CBF \cite{lopez2020robust}, Gaussian Process CBF\cite{castaneda2021gaussian}, Gaussiance Process robust adaptive CBF\cite{ykgpracbf}, disturbance observer-based CBF \cite{Wang2023}\cite{Ersin2022}, and several neural network-based CBFs \cite{ChengRLCBF} \cite{WangCDC2022} \cite{manda2024domainadaptivesafetyfilters}.

Lastly, tuning the $\alpha$ value is non-trivial, as the parameter depends on model uncertainty and the size of the safe region, while significantly affecting the performances of CBF-based controllers. For example, when the safe region is very small and there exists uncertainty in the system model, CBF-based controllers might not be able to find the feasible safe control input unless $\alpha$ is not properly selected. To address this issue, we can use the sampling-based approach for tuning $\alpha$ proposed by \cite{10644811}. First, we sample states that satisfy the hard safe constraint condition, for example, $\dot{h}\geq \alpha(h)$. Using the sampled states, we can construct the approximated safe set and consequently verify whether the selected $\alpha$ is valid. We expect that an iterative tuning algorithm for $\alpha$ can be incorporated to enhance the applicability of the proposed framework in real-world experiments in future research.

%% file: sections/conclusions.tex
\section{Conclusion and Future Work}\label{sec:conclusion}
This study introduced a novel framework that ensures the safety of ESD automation in a theoretically rigorous way. By employing CBFs, the proposed framework enables the robot to accurately respect the boundaries of individual tumors, even when they are located close to one another within the GI tract. This capability ensures the targeted treatment and removal of each tumor while protecting the surrounding normal mucosal tissue.

In future work, the proposed framework will be further developed to move beyond simulations, enabling its application to realistic phantom models or ex-vivo environments. In particular, the focus will be on developing a method to estimate the linear velocity of the end-effector in a tendon-driven robot based on motor rotational speed, utilizing the robot's kinematics and a nonlinear hysteresis model for real-world applications of the proposed control framework. Furthermore, through the integration of the endoscopic camera and deep learning algorithms, it will be possible to acquire geometric information of tumors in the task environment.

%% file: sections/appendices.tex
\appendices

\bibliographystyle{IEEEtran}
\bibliography{bibliography}

\section{Author Information} 
\begin{IEEEbiography}[{\includegraphics[width=1in,height=1.25in,clip,keepaspectratio]{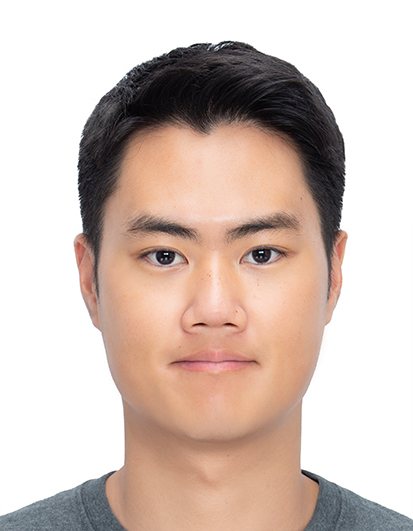}}]{Yitaek Kim} received the B.S. degree from the Division of Robotics, Kwangwoon University, Seoul, Republic of Korea, in 2017, M.S. degree in Interdisciplinary Engineering Systems from Hanyang University in 2019, and Ph.D. degree at the Mærsk McKinney Møller Institute, University of Southern Denmark, Odense, Denmark. 

From September 2023 to December 2023, he was a visiting student researcher at AMBER Lab in the Department of Mechanical and Civil Engineering, California Institute of Technology (Caltech), Pasadena, USA. He is currently Post-doc with the Mærsk McKinney Møller Institute, University of Southern Denmark, Odense, Denmark. His research interests revolve around robot control and safety-critical control systems.
\end{IEEEbiography}

\begin{IEEEbiography}[{\includegraphics[width=1in,height=1.25in,clip,keepaspectratio]{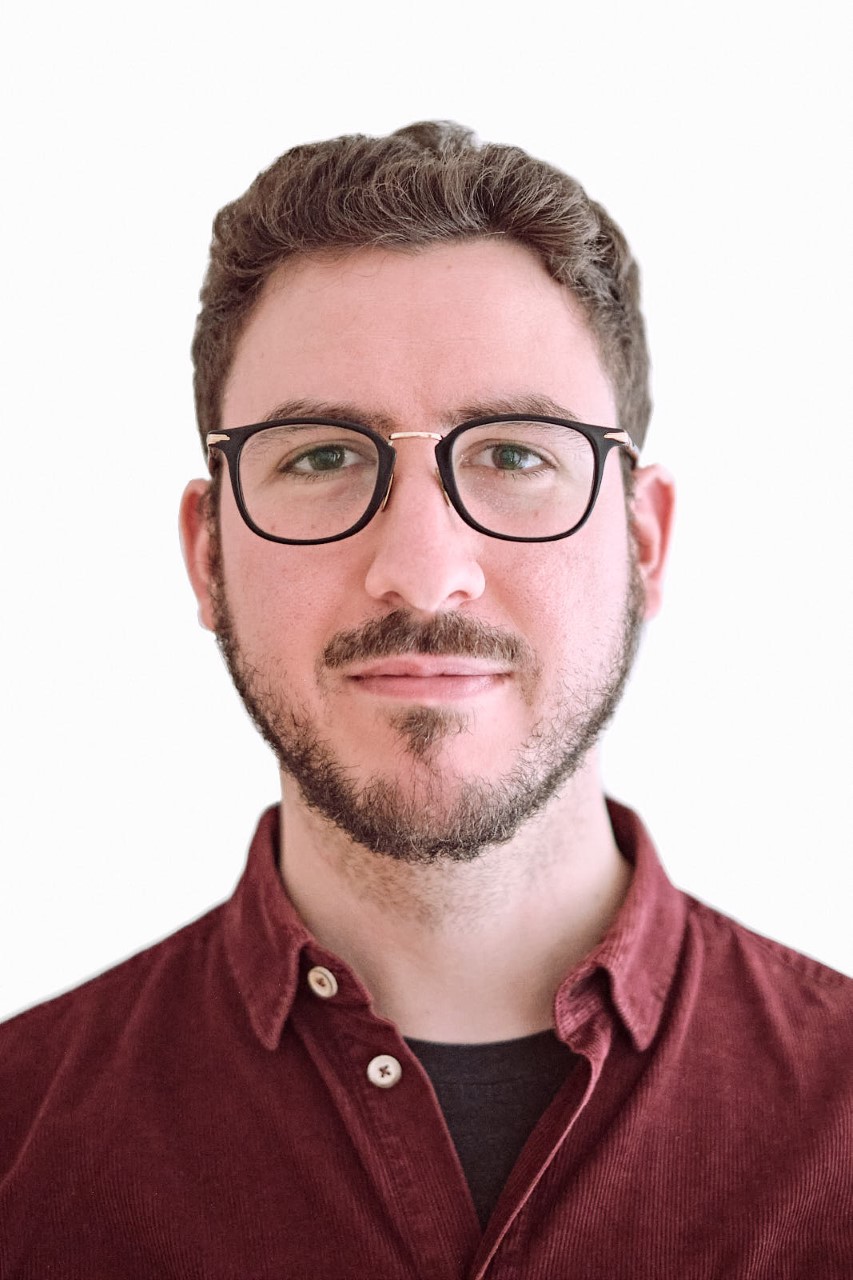}}]
{Iñigo Iturrate} received the M.Sc. degree in robot systems engineering from the University of Southern Denmark, in 2015, and industrial Ph.D. degree from the Mærsk McKinney Møller Institute, University of Southern Denmark, in collaboration with Universal Robots A/S in 2019.

He is currently Associate Professor with the Mærsk McKinney Møller Institute, University of Southern Denmark, Odense, Denmark. His research interests include robot imitation learning, uncertainty quantification in robot policy learning, artificial intelligence, and safety-critical and adaptive control, with applications to industrial (re)manufacturing, robotic surgery and medical diagnostics.
\end{IEEEbiography}

\begin{IEEEbiography}[{\includegraphics[width=1in,height=1.25in,clip,keepaspectratio]{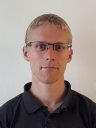}}]
{Christoffer Sloth} received the M.Sc. degree in control engineering from Aalborg University, Denmark in 2009, and Ph.D. degree in computer science from Aalborg University, Denmark in 2012. 

From 2016 to 2017, he was Associate Professor at Section for Automation and Control at Aalborg University, and since 2017 he has been employed at the Mærsk McKinney Møller Institute, University of Southern Denmark, Odense, Denmark where he is currently Professor. Christoffer Sloth does research within control, estimation and optimization with application to robotics.
\end{IEEEbiography}

\begin{IEEEbiography}[{\includegraphics[width=1in,height=1.25in,clip,keepaspectratio]{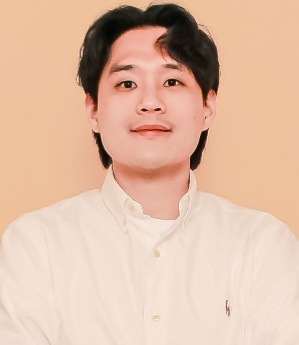}}]{Hansoul Kim} received the B.S. degree from the Division of Robotics, Kwangwoon University, Seoul, Republic of Korea, in 2017, the M.S. degree from the Robotics Program, Korea Advanced Institute of Science and Technology (KAIST), Daejeon, Republic of Korea, in 2019, and the Ph.D. degree from the Department of Mechanical Engineering, KAIST, in 2022.

From 2022 to 2023, he served as a Post-doctoral researcher at the Walker Department of Mechanical Engineering and Texas Robotics, the University of Texas at Austin, TX, USA. From 2023 to 2024, he served as as a Post-doctoral researcher in the Department of Electrical Engineering and Computer Sciences at the University of California, Berkeley, CA, USA. He is currently an Assistant Professor with the Division of Mechanical System Engineering at the Myongji University, Yongin, Republic of Korea. His research interests primarily focus on surgical robotics, AI-assisted robot surgery, robot task automation, soft robotics, and mechanism design.

\end{IEEEbiography}





\EOD

%% file: access.bbl
\begin{thebibliography}{10}
\providecommand{\url}[1]{#1}
\csname url@samestyle\endcsname
\providecommand{\newblock}{\relax}
\providecommand{\bibinfo}[2]{#2}
\providecommand{\BIBentrySTDinterwordspacing}{\spaceskip=0pt\relax}
\providecommand{\BIBentryALTinterwordstretchfactor}{4}
\providecommand{\BIBentryALTinterwordspacing}{\spaceskip=\fontdimen2\font plus
\BIBentryALTinterwordstretchfactor\fontdimen3\font minus \fontdimen4\font\relax}
\providecommand{\BIBforeignlanguage}[2]{{%
\expandafter\ifx\csname l@#1\endcsname\relax
\typeout{** WARNING: IEEEtran.bst: No hyphenation pattern has been}%
\typeout{** loaded for the language `#1'. Using the pattern for}%
\typeout{** the default language instead.}%
\else
\language=\csname l@#1\endcsname
\fi
#2}}
\providecommand{\BIBdecl}{\relax}
\BIBdecl

\bibitem{bray2018global}
F.~Bray, J.~Ferlay, I.~Soerjomataram, R.~L. Siegel, L.~A. Torre, and A.~Jemal, ``Global cancer statistics 2018: Globocan estimates of incidence and mortality worldwide for 36 cancers in 185 countries,'' \emph{CA: a cancer journal for clinicians}, vol.~68, no.~6, pp. 394--424, 2018.

\bibitem{hamashima2015mortality}
C.~Hamashima, M.~Shabana, K.~Okada, M.~Okamoto, and Y.~Osaki, ``Mortality reduction from gastric cancer by endoscopic and radiographic screening,'' \emph{Cancer science}, vol. 106, no.~12, pp. 1744--1749, 2015.

\bibitem{dekker2018advances}
E.~Dekker and D.~K. Rex, ``Advances in {CRC} prevention: screening and surveillance,'' \emph{Gastroenterology}, vol. 154, no.~7, pp. 1970--1984, 2018.

\bibitem{bevan2018colorectal}
R.~Bevan, M.~D. Rutter \emph{et~al.}, ``Colorectal cancer screening-who, how, and when,'' \emph{Clin Endosc}, vol.~51, no.~1, pp. 37--49, 2018.

\bibitem{schreuders2016advances}
E.~H. Schreuders, E.~J. Grobbee, M.~C. Spaander, and E.~J. Kuipers, ``Advances in fecal tests for colorectal cancer screening,'' \emph{Current treatment options in gastroenterology}, vol.~14, pp. 152--162, 2016.

\bibitem{lee2010two}
S.~H. Lee, J.~E. Kwon, and Y.~S. Cheong, ``Two cases of trichuris trichiura infection diagnosed by colonoscopy.'' \emph{Korean Journal of Family Medicine}, vol.~31, no.~8, pp. 622--629, 2010.

\bibitem{lee2019robotic}
D.-H. Lee, M.~Hwang, and D.-S. Kwon, ``Robotic endoscopy system (easyendo) with a robotic arm mountable on a conventional endoscope,'' in \emph{2019 International Conference on Robotics and Automation (ICRA)}.\hskip 1em plus 0.5em minus 0.4em\relax IEEE, 2019, pp. 367--372.

\bibitem{ruiter2012design}
J.~Ruiter, E.~Rozeboom, M.~van~der Voort, M.~Bonnema, and I.~Broeders, ``Design and evaluation of robotic steering of a flexible endoscope,'' in \emph{2012 4th IEEE RAS \& EMBS international conference on biomedical robotics and biomechatronics (BioRob)}.\hskip 1em plus 0.5em minus 0.4em\relax IEEE, 2012, pp. 761--767.

\bibitem{abbott2007design}
D.~J. Abbott, C.~Becke, R.~I. Rothstein, and W.~J. Peine, ``Design of an endoluminal notes robotic system,'' in \emph{2007 IEEE/RSJ International Conference on Intelligent Robots and Systems}.\hskip 1em plus 0.5em minus 0.4em\relax IEEE, 2007, pp. 410--416.

\bibitem{lomanto2015flexible}
D.~Lomanto, S.~Wijerathne, L.~K.~Y. Ho, and L.~S.~J. Phee, ``Flexible endoscopic robot,'' \emph{Minimally Invasive Therapy \& Allied Technologies}, vol.~24, no.~1, pp. 37--44, 2015.

\bibitem{chiu2021colonic}
P.~W.~Y. Chiu, K.~Y. Ho, and S.~J. Phee, ``Colonic endoscopic submucosal dissection using a novel robotic system (with video),'' \emph{Gastrointestinal endoscopy}, vol.~93, no.~5, pp. 1172--1177, 2021.

\bibitem{hwang2020evaluation}
M.~Hwang, S.~W. Lee, K.~C. Park, H.~J. Sul, and D.-S. Kwon, ``Evaluation of a robotic arm-assisted endoscope to facilitate endoscopic submucosal dissection (with video),'' \emph{Gastrointestinal Endoscopy}, vol.~91, no.~3, pp. 699--706, 2020.

\bibitem{berthet20182}
P.~Berthet-Rayne, G.~Gras, K.~Leibrandt, P.~Wisanuvej, A.~Schmitz, C.~A. Seneci, and G.-Z. Yang, ``The {i2Snake} robotic platform for endoscopic surgery,'' \emph{Annals of biomedical engineering}, vol.~46, pp. 1663--1675, 2018.

\bibitem{nageotte2020stras}
F.~Nageotte, L.~Zorn, P.~Zanne, and M.~De~Mathelin, ``Stras: a modular and flexible telemanipulated robotic device for intraluminal surgery,'' in \emph{Handbook of robotic and image-guided surgery}.\hskip 1em plus 0.5em minus 0.4em\relax Elsevier, 2020, pp. 123--146.

\bibitem{nakadate2020surgical}
R.~Nakadate, T.~Iwasa, S.~Onogi, J.~Arata, S.~Oguri, Y.~Okamoto, T.~Akahoshi, M.~Eto, and M.~Hashizume, ``Surgical robot for intraluminal access: An ex vivo feasibility study,'' \emph{Cyborg and Bionic Systems}, 2020.

\bibitem{hwang2020k}
M.~Hwang and D.-S. Kwon, ``K-flex: a flexible robotic platform for scar-free endoscopic surgery,'' \emph{The International Journal of Medical Robotics and Computer Assisted Surgery}, vol.~16, no.~2, p. e2078, 2020.

\bibitem{kim2023endoscopic}
H.~Kim, J.~M. You, K.-U. Kyung, and D.-S. Kwon, ``Endoscopic surgery robot that facilitates insertion of the curved colon and ensures positional stability against external forces: K-colon,'' \emph{The International Journal of Medical Robotics and Computer Assisted Surgery}, vol.~19, no.~3, p. e2493, 2023.

\bibitem{hwang2022automating}
M.~Hwang, J.~Ichnowski, B.~Thananjeyan, D.~Seita, S.~Paradis, D.~Fer, T.~Low, and K.~Goldberg, ``Automating surgical peg transfer: Calibration with deep learning can exceed speed, accuracy, and consistency of humans,'' \emph{IEEE Transactions on Automation Science and Engineering}, vol.~20, no.~2, pp. 909--922, 2022.

\bibitem{hwang2020superhuman}
M.~Hwang, B.~Thananjeyan, D.~Seita, J.~Ichnowski, S.~Paradis, D.~Fer, T.~Low, and K.~Goldberg, ``Superhuman surgical peg transfer using depth-sensing and deep recurrent neural networks,'' \emph{arXiv preprint arXiv:2012.12844}, 2020.

\bibitem{rosen2015autonomous}
J.~Rosen and J.~Ma, ``Autonomous operation in surgical robotics,'' \emph{Mechanical Engineering}, vol. 137, no.~09, pp. S15--S18, 2015.

\bibitem{lum2008telerobotic}
M.~J. Lum, J.~Rosen, T.~S. Lendvay, A.~S. Wright, M.~N. Sinanan, and B.~Hannaford, ``Telerobotic fundamentals of laparoscopic surgery (fls): effects of time delay-pilot study,'' in \emph{2008 30th Annual International Conference of the IEEE Engineering in Medicine and Biology Society}.\hskip 1em plus 0.5em minus 0.4em\relax IEEE, 2008, pp. 5597--5600.

\bibitem{gonzalez2021deserts}
G.~Gonzalez, M.~Agarwal, M.~V. Balakuntala, M.~M. Rahman, U.~Kaur, R.~M. Voyles, V.~Aggarwal, Y.~Xue, and J.~Wachs, ``Deserts: Delay-tolerant semi-autonomous robot teleoperation for surgery,'' in \emph{2021 IEEE International Conference on Robotics and Automation (ICRA)}.\hskip 1em plus 0.5em minus 0.4em\relax IEEE, 2021, pp. 12\,693--12\,700.

\bibitem{xu2021surrol}
J.~Xu, B.~Li, B.~Lu, Y.-H. Liu, Q.~Dou, and P.-A. Heng, ``Surrol: An open-source reinforcement learning centered and dvrk compatible platform for surgical robot learning,'' in \emph{2021 IEEE/RSJ International Conference on Intelligent Robots and Systems (IROS)}.\hskip 1em plus 0.5em minus 0.4em\relax IEEE, 2021, pp. 1821--1828.

\bibitem{seita2018fast}
D.~Seita, S.~Krishnan, R.~Fox, S.~McKinley, J.~Canny, and K.~Goldberg, ``Fast and reliable autonomous surgical debridement with cable-driven robots using a two-phase calibration procedure,'' in \emph{2018 IEEE International Conference on Robotics and Automation (ICRA)}.\hskip 1em plus 0.5em minus 0.4em\relax IEEE, 2018, pp. 6651--6658.

\bibitem{murali2015learning}
A.~Murali, S.~Sen, B.~Kehoe, A.~Garg, S.~McFarland, S.~Patil, W.~D. Boyd, S.~Lim, P.~Abbeel, and K.~Goldberg, ``Learning by observation for surgical subtasks: Multilateral cutting of 3d viscoelastic and 2d orthotropic tissue phantoms,'' in \emph{2015 IEEE International Conference on Robotics and Automation (ICRA)}.\hskip 1em plus 0.5em minus 0.4em\relax IEEE, 2015, pp. 1202--1209.

\bibitem{dharmarajan2023automating}
K.~Dharmarajan, W.~Panitch, M.~Jiang, K.~Srinivas, B.~Shi, Y.~Avigal, H.~Huang, T.~Low, D.~Fer, and K.~Goldberg, ``Automating vascular shunt insertion with the dvrk surgical robot,'' in \emph{2023 IEEE International Conference on Robotics and Automation (ICRA)}.\hskip 1em plus 0.5em minus 0.4em\relax IEEE, 2023, pp. 6781--6788.

\bibitem{dharmarajan2023trimodal}
K.~Dharmarajan, W.~Panitch, B.~Shi, H.~Huang, L.~Y. Chen, T.~Low, D.~Fer, and K.~Goldberg, ``A trimodal framework for robot-assisted vascular shunt insertion when a supervising surgeon is local, remote, or unavailable,'' in \emph{2023 International Symposium on Medical Robotics (ISMR)}.\hskip 1em plus 0.5em minus 0.4em\relax IEEE, 2023, pp. 1--8.

\bibitem{dharmarajan2023robot}
K.~Dharmarajan, W.~Panitch, B.~Shi, H.~Huang, L.~Y. Chen, M.~Moghani, Q.~Yu, K.~Hari, T.~Low, D.~Fer \emph{et~al.}, ``Robot-assisted vascular shunt insertion with the dvrk surgical robot,'' \emph{Journal of Medical Robotics Research}, vol.~8, no. 03n04, p. 2340006, 2023.

\bibitem{ferraguti2022safety}
F.~Ferraguti, C.~T. Landi, A.~Singletary, H.-C. Lin, A.~Ames, C.~Secchi, and M.~Bonf{\`e}, ``Safety and efficiency in robotics: The control barrier functions approach,'' \emph{IEEE Robotics \& Automation Magazine}, vol.~29, no.~3, pp. 139--151, 2022.

\bibitem{bowyer2013active}
S.~A. Bowyer, B.~L. Davies, and F.~R. y~Baena, ``Active constraints/virtual fixtures: A survey,'' \emph{IEEE Transactions on Robotics}, vol.~30, no.~1, pp. 138--157, 2013.

\bibitem{attanasio2021autonomy}
A.~Attanasio, B.~Scaglioni, E.~De~Momi, P.~Fiorini, and P.~Valdastri, ``Autonomy in surgical robotics,'' \emph{Annual Review of Control, Robotics, and Autonomous Systems}, vol.~4, no.~1, pp. 651--679, 2021.

\bibitem{Ames2019CBFtheoryandapplications}
A.~D. Ames, S.~Coogan, M.~Egerstedt, G.~Notomista, K.~Sreenath, and P.~Tabuada, ``Control barrier functions: Theory and applications,'' in \emph{18th European Control Conference (ECC)}, 2019, pp. 3420--3431.

\bibitem{WOLF202218}
Ádám Wolf, D.~Wolton, J.~Trapl, J.~Janda, S.~Romeder-Finger, T.~Gatternig, J.-B. Farcet, P.~Galambos, and K.~Széll, ``Towards robotic laboratory automation plug play: The “lapp” framework,'' \emph{SLAS Technology}, vol.~27, no.~1, pp. 18--25, 2022.

\bibitem{Hadi2024}
H.~Jahanshahi and Z.~H. Zhu, ``Review of machine learning in robotic grasping control in space application,'' \emph{Acta Astronautica}, vol. 220, pp. 37--61, 2024.

\bibitem{YKECC2024}
Y.~Kim, J.~Kim, A.~D. Ames, and C.~Sloth, ``Robust safety-critical control for input-delayed system with delay estimation,'' in \emph{2024 European Control Conference (ECC)}, 2024, pp. 2218--2223.

\bibitem{lopez2020robust}
B.~T. Lopez, J.-J.~E. Slotine, and J.~P. How, ``Robust adaptive control barrier functions: An adaptive and data-driven approach to safety,'' \emph{IEEE Control Systems Letters}, vol.~5, no.~3, pp. 1031--1036, 2020.

\bibitem{JANKOVIC2018359}
M.~Jankovic, ``Robust control barrier functions for constrained stabilization of nonlinear systems,'' \emph{Automatica}, vol.~96, pp. 359--367, 2018.

\bibitem{ykgpracbf}
Y.~Kim, I.~Iturrate, J.~Langaa, and C.~Sloth, ``Safe robust adaptive control under both parametric and nonparametric uncertainty,'' \emph{Advanced Robotics}, pp. 1--10, 2024.

\bibitem{Castaneda2021GPCBF}
F.~Castañeda, J.~J. Choi, B.~Zhang, C.~J. Tomlin, and K.~Sreenath, ``Pointwise feasibility of {G}aussian process-based safety-critical control under model uncertainty,'' in \emph{60th IEEE Conference on Decision and Control (CDC)}, 2021, pp. 6762--6769.

\bibitem{molnar2021model}
T.~G. Molnar, R.~K. Cosner, A.~W. Singletary, W.~Ubellacker, and A.~D. Ames, ``Model-free safety-critical control for robotic systems,'' \emph{IEEE robotics and automation letters}, vol.~7, no.~2, pp. 944--951, 2021.

\bibitem{Spong2005}
M.~W. Spong, S.~Hutchinson, and M.~Vidyasagar, \emph{Robot Modeling and Control}.\hskip 1em plus 0.5em minus 0.4em\relax New York: John Wiley and Sons, 2005.

\bibitem{oyama2005endoscopic}
T.~Oyama, A.~Tomori, K.~Hotta, S.~Morita, K.~Kominato, M.~Tanaka, and Y.~Miyata, ``Endoscopic submucosal dissection of early esophageal cancer,'' \emph{Clinical Gastroenterology and Hepatology}, vol.~3, no.~7, pp. S67--S70, 2005.

\bibitem{sontagISS2008}
E.~D. Sontag, ``Input to state stability: Basic concepts and results,'' in \emph{Nonlinear and Optimal Control Theory}.\hskip 1em plus 0.5em minus 0.4em\relax Springer, 2008, pp. 163--220.

\bibitem{ShishirISSfCBFs2019}
S.~Kolathaya and A.~D. Ames, ``Input-to-state safety with control barrier functions,'' \emph{IEEE Control Systems Letters}, vol.~3, no.~1, pp. 108--113, 2019.

\bibitem{Matni2024}
N.~Matni, A.~D. Ames, and J.~C. Doyle, ``A quantitative framework for layered multirate control: Toward a theory of control architecture,'' \emph{IEEE Control Systems Magazine}, vol.~44, no.~3, pp. 52--94, 2024.

\bibitem{lambin2021endoscopic}
T.~Lambin, J.~Rivory, T.~Wallenhorst, R.~Legros, F.~Monzy, J.~Jacques, and M.~Pioche, ``Endoscopic submucosal dissection: How to be more efficient?'' \emph{Endoscopy International Open}, vol.~9, no.~11, pp. E1720--E1730, 2021.

\bibitem{kim2024efficacy}
J.~Kim, D.-H. Lee, D.-S. Kwon, K.~C. Park, H.~J. Sul, M.~Hwang, and S.-W. Lee, ``Efficacy of robot arm-assisted endoscopic submucosal dissection in live porcine stomach (with video),'' \emph{Scientific Reports}, vol.~14, no.~1, p. 17367, 2024.

\bibitem{dabbene2017simple}
F.~Dabbene, D.~Henrion, and C.~M. Lagoa, ``Simple approximations of semialgebraic sets and their applications to control,'' \emph{Automatica}, vol.~78, pp. 110--118, 2017.

\bibitem{kim2020effect}
H.~Kim, M.~Hwang, J.~Kim, J.~M. You, C.-S. Lim, and D.-S. Kwon, ``Effect of backlash hysteresis of surgical tool bending joints on task performance in teleoperated flexible endoscopic robot,'' \emph{The International Journal of Medical Robotics and Computer Assisted Surgery}, vol.~16, no.~1, p. e2047, 2020.

\bibitem{do2014hysteresis}
T.~Do, T.~Tjahjowidodo, M.~Lau, T.~Yamamoto, and S.~Phee, ``Hysteresis modeling and position control of tendon-sheath mechanism in flexible endoscopic systems,'' \emph{Mechatronics}, vol.~24, no.~1, pp. 12--22, 2014.

\bibitem{le2016survey}
H.~M. Le, T.~N. Do, and S.~J. Phee, ``A survey on actuators-driven surgical robots,'' \emph{Sensors and Actuators A: Physical}, vol. 247, pp. 323--354, 2016.

\bibitem{lipson2021raft}
L.~Lipson, Z.~Teed, and J.~Deng, ``Raft-stereo: Multilevel recurrent field transforms for stereo matching,'' in \emph{2021 International Conference on 3D Vision (3DV)}.\hskip 1em plus 0.5em minus 0.4em\relax IEEE, 2021, pp. 218--227.

\bibitem{ma2024segment}
J.~Ma, Y.~He, F.~Li, L.~Han, C.~You, and B.~Wang, ``Segment anything in medical images,'' \emph{Nature Communications}, vol.~15, no.~1, p. 654, 2024.

\bibitem{pmlr-v155-boffi21a}
\BIBentryALTinterwordspacing
N.~Boffi, S.~Tu, N.~Matni, J.-J. Slotine, and V.~Sindhwani, ``Learning stability certificates from data,'' in \emph{Proceedings of the 2020 Conference on Robot Learning}, ser. Proceedings of Machine Learning Research, J.~Kober, F.~Ramos, and C.~Tomlin, Eds., vol. 155.\hskip 1em plus 0.5em minus 0.4em\relax PMLR, 16--18 Nov 2021, pp. 1341--1350. [Online]. Available: \url{https://proceedings.mlr.press/v155/boffi21a.html}
\BIBentrySTDinterwordspacing

\bibitem{Taylor2020}
A.~J. Taylor and A.~D. Ames, ``Adaptive safety with control barrier functions,'' in \emph{2020 American Control Conference (ACC)}, 2020, pp. 1399--1405.

\bibitem{castaneda2021gaussian}
F.~Casta{\~n}eda, J.~J. Choi, B.~Zhang, C.~J. Tomlin, and K.~Sreenath, ``Gaussian process-based min-norm stabilizing controller for control-affine systems with uncertain input effects and dynamics,'' in \emph{2021 American Control Conference (ACC)}, 2021, pp. 3683--3690.

\bibitem{Wang2023}
Y.~Wang and X.~Xu, ``Disturbance observer-based robust control barrier functions,'' in \emph{2023 American Control Conference (ACC)}, 2023, pp. 3681--3687.

\bibitem{Ersin2022}
E.~Daş and R.~M. Murray, ``Robust safe control synthesis with disturbance observer-based control barrier functions,'' in \emph{2022 IEEE 61st Conference on Decision and Control (CDC)}, 2022, pp. 5566--5573.

\bibitem{ChengRLCBF}
\BIBentryALTinterwordspacing
R.~Cheng, G.~Orosz, R.~M. Murray, and J.~W. Burdick, ``End-to-end safe reinforcement learning through barrier functions for safety-critical continuous control tasks,'' in \emph{Proceedings of the Thirty-Third AAAI Conference on Artificial Intelligence and Thirty-First Innovative Applications of Artificial Intelligence Conference and Ninth AAAI Symposium on Educational Advances in Artificial Intelligence}, ser. AAAI'19/IAAI'19/EAAI'19.\hskip 1em plus 0.5em minus 0.4em\relax AAAI Press, 2019. [Online]. Available: \url{https://doi.org/10.1609/aaai.v33i01.33013387}
\BIBentrySTDinterwordspacing

\bibitem{WangCDC2022}
C.~Wang, Y.~Meng, S.~L. Smith, and J.~Liu, ``Data-driven learning of safety-critical control with stochastic control barrier functions,'' in \emph{2022 IEEE 61st Conference on Decision and Control (CDC)}, 2022, pp. 5309--5315.

\bibitem{manda2024domainadaptivesafetyfilters}
\BIBentryALTinterwordspacing
L.~Manda, S.~Chen, and M.~Fazlyab, ``Domain adaptive safety filters via deep operator learning,'' 2024. [Online]. Available: \url{https://arxiv.org/abs/2410.14528}
\BIBentrySTDinterwordspacing

\bibitem{10644811}
J.~Lee, J.~Kim, and A.~D. Ames, ``A data-driven method for safety-critical control: Designing control barrier functions from state constraints,'' in \emph{2024 American Control Conference (ACC)}, 2024, pp. 394--401.

\end{thebibliography}
